\documentclass{article}

\usepackage[final]{neurips_2022}

\usepackage[utf8]{inputenc} 
\usepackage[T1]{fontenc}    
\usepackage[hidelinks]{hyperref}       
\usepackage{url}            
\usepackage{booktabs}       
\usepackage{amsfonts}       
\usepackage{nicefrac}       
\usepackage{microtype}      
\usepackage{amsmath}
\usepackage{svg}
\usepackage{wrapfig}
\usepackage{amssymb}
\usepackage{graphicx}
\usepackage{xcolor}
\usepackage{subcaption}
\usepackage{caption}
\usepackage{hyperref}
\usepackage{array}
\usepackage{multirow}
\usepackage{tablefootnote}
\usepackage{attrib}

\usepackage{pifont}

\usepackage{mathtools}
\usepackage{xcolor,colortbl}

\DeclarePairedDelimiter{\norm}{\lVert}{\rVert}

\title{\texttt{LLM.int8()}: 8-bit Matrix Multiplication \\ for Transformers at Scale}

\author{Tim Dettmers$^{\lambda}$\thanks{Majority of research done as a visiting researcher at Facebook AI Research.} \And Mike Lewis$^\dagger$ \And Younes Belkada$^{\mathsection\mp}$ \\\And Luke Zettlemoyer$^{\dagger\lambda}$
  \AND \\ University of Washington$^\lambda$ \\Facebook AI Research$^\dagger$\\ Hugging Face$^\mathsection$\\ ENS Paris-Saclay$^\mp$
}

\begin{document}

\maketitle

\begin{abstract}
Large language models have been widely adopted but require significant GPU memory for inference. We develop a procedure for Int8 matrix multiplication for feed-forward and attention projection layers in transformers, which cut the memory needed for inference by half while retaining full precision performance. With our method, a 175B parameter 16/32-bit checkpoint can be loaded, converted to Int8, and used immediately without performance degradation. This is made possible by understanding and working around properties of highly systematic emergent features in transformer language models that dominate attention and transformer predictive performance. To cope with these features, we develop a two-part quantization procedure, {\bf LLM.int8()}. We first use vector-wise quantization with separate normalization constants for each inner product in the matrix multiplication, to quantize most of the features. However, for the emergent outliers, we also include a new mixed-precision decomposition scheme, which isolates the outlier feature dimensions into a 16-bit matrix multiplication while still more than 99.9\% of values are multiplied in 8-bit. Using LLM.int8(), we show empirically it is possible to perform inference in LLMs with up to 175B parameters without any performance degradation. This result makes such models much more accessible, for example making it possible to use OPT-175B/BLOOM on a single server with consumer GPUs.
We open source our software.

\end{abstract}

\section{Introduction}

Large pretrained language models are widely adopted in NLP \citep{vaswani2017attention,radford2019language, brown2020language, zhang2022opt} but require significant memory for inference. For large transformer language models at and beyond 6.7B parameters, the feed-forward and attention projection layers and their matrix multiplication operations are responsible for 95\%\footnote{Other parameters come mostly from the embedding layer. A tiny amount comes from norms and biases.} of consumed parameters and 65-85\% of all computation \citep{ilharco-etal-2020-high}. One way to reduce the size of the parameters is to quantize them to less bits and use low-bit-precision matrix multiplication. With this goal in mind, 8-bit quantization methods for transformers have been developed \citep{chen2020statistical, lin2020towards, zafrir2019q8bert,shen2020q}. 
While these methods reduce memory use, they degrade performance, usually require tuning quantization further after training, and have only been studied for models with less than 350M parameters. Degradation-free quantization up to 350M parameters is poorly understood, and multi-billion parameter quantization remains an open challenge.

\begin{wrapfigure}{r}{0.5\textwidth}
\centering
         \includegraphics[scale=0.30]{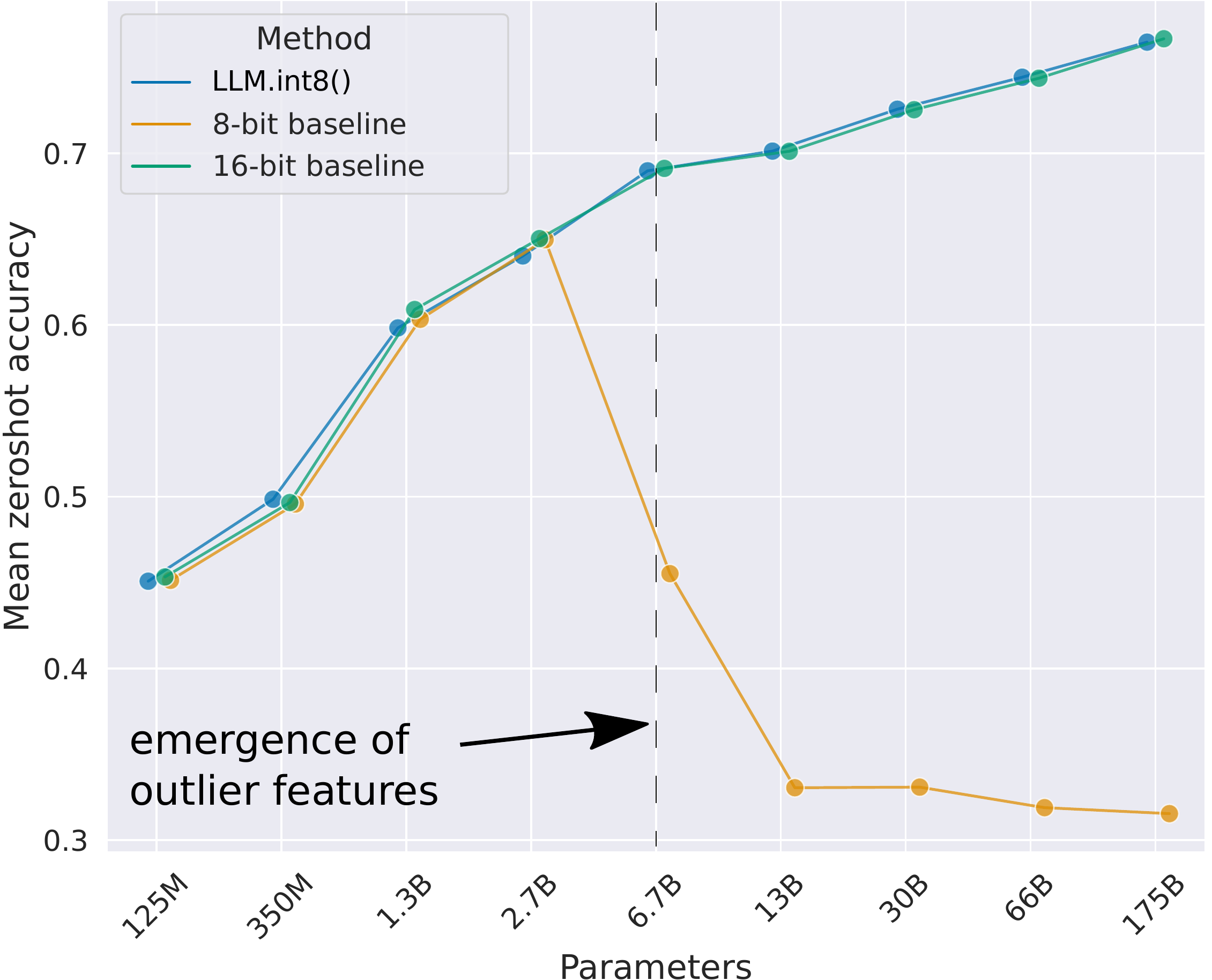}
        \caption{OPT model mean zeroshot accuracy for WinoGrande, HellaSwag, PIQA, and LAMBADA datasets. Shown is the 16-bit baseline, the most precise previous 8-bit quantization method as a baseline, and our new 8-bit quantization method, LLM.int8(). We can see once systematic outliers occur at a scale of 6.7B parameters, regular quantization methods fail, while LLM.int8() maintains 16-bit accuracy.}
        \label{fig:results}
\end{wrapfigure}

In this paper, we present the first multi-billion-scale Int8 quantization procedure for transformers that does not incur any performance degradation. Our procedure makes it possible to load a 175B parameter transformer with 16 or 32-bit weights, convert the feed-forward and attention projection layers to 8-bit, and use the resulting model immediately for inference without any performance degradation.
We achieve this result by solving two key challenges: the need for higher quantization precision at scales beyond 1B parameters and the need to explicitly represent the sparse but systematic large magnitude outlier features that ruin quantization precision once they emerge in {\it all} transformer layers starting at scales of 6.7B parameters. This loss of precision is reflected in C4 evaluation perplexity (Section~\ref{sec:matmulatscale}) as well as zeroshot accuracy as soon as these outlier features emerge, as shown in Figure~\ref{fig:results}.

We show that with the first part of our method, vector-wise quantization, it is possible to retain performance at scales up to 2.7B parameters. For vector-wise quantization,  matrix multiplication can be seen as a sequence of independent inner products of row and column vectors. As such, we can use a separate quantization normalization constant for each inner product to improve quantization precision. We can recover the output of the matrix multiplication by denormalizing by the outer product of column and row normalization constants before we perform the next operation. 

To scale beyond 6.7B parameters without performance degradation, it is critical to understand the emergence of extreme outliers in the feature dimensions of the hidden states during inference. To this end, we provide a new descriptive analysis which shows that large features with magnitudes up to 20x larger than in other dimensions first appear in about 25\% of all transformer layers and then gradually spread to other layers as we scale transformers to 6B parameters. At around 6.7B parameters, a phase shift occurs, and {\it all} transformer layers and 75\% of all sequence dimensions are affected by extreme magnitude features. These outliers are highly systematic: at the 6.7B scale, 150,000 outliers occur per sequence, but they are concentrated in only 6 feature dimensions across the entire transformer. Setting these outlier feature dimensions to zero decreases top-1 attention softmax probability mass by more than 20\% and degrades validation perplexity by 600-1000\% despite them only making up about 0.1\% of all input features. In contrast, removing the same amount of random features decreases the probability by a maximum of 0.3\% and degrades perplexity by about 0.1\%.

To support effective quantization with such extreme outliers, we develop mixed-precision decomposition, the second part of our method. We perform 16-bit matrix multiplication for the outlier feature dimensions and 8-bit matrix multiplication for the other 99.9\% of the dimensions. We name the combination of vector-wise quantization and mixed precision decomposition, {\bf LLM.int8()}. We show that by using LLM.int8(), we can perform inference in LLMs with up to 175B parameters without any performance degradation. Our method not only provides new insights into the effects of these outliers on model performance but also makes
it possible for the first time to use very large models, for example, OPT-175B/BLOOM, on a single server with consumer GPUs. While our work focuses on making large language models accessible without degradation, we also show in Appendix~\ref{app:inference} that we maintain end-to-end inference runtime performance for large models, such as BLOOM-176B and provide modest matrix multiplication speedups for GPT-3 models of size 6.7B parameters or larger. We open-source our software\footnote{\url{https://github.com/TimDettmers/bitsandbytes}} and release a Hugging Face Transformers \citep{wolf2019huggingface} integration making our method available to all hosted Hugging Face Models that have linear layers.

\begin{figure}[t]
     \centering
         \includegraphics[scale=0.12]{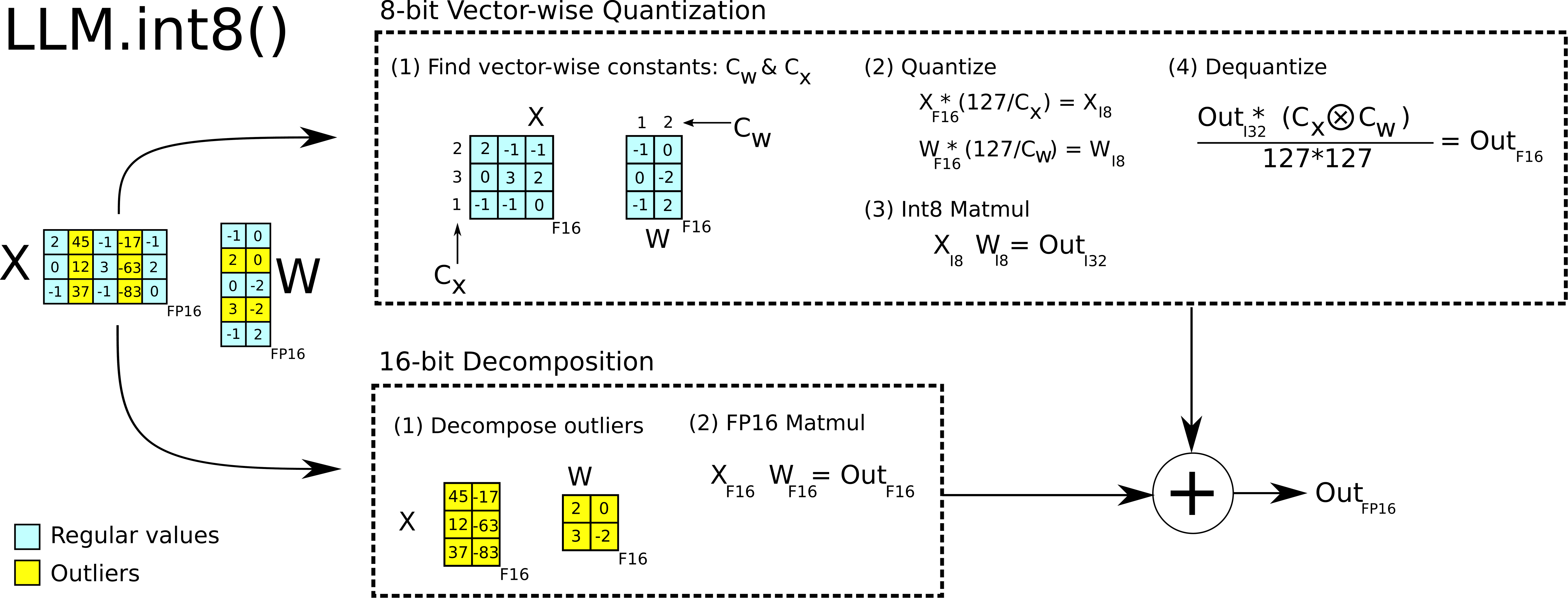}
        \caption{Schematic of LLM.int8(). Given 16-bit floating-point inputs $\mathbf{X}_{f16}$ and weights $\mathbf{W}_{f16}$, the features and weights are decomposed into sub-matrices of large magnitude features and other values. The outlier feature matrices are multiplied in 16-bit. All other values are multiplied in 8-bit. We perform 8-bit vector-wise multiplication by scaling by row and column-wise absolute maximum of $\mathbf{C}_x$ and $\mathbf{C}_w$ and then quantizing the outputs to Int8. The Int32 matrix multiplication outputs $\mathbf{{Out}}_{i32}$ are dequantization by the outer product of the normalization constants $\mathbf{C}_x\otimes \mathbf{C}_w$. Finally, both outlier and regular outputs are accumulated in 16-bit floating point outputs.}
        \label{fig:method}
\end{figure}

\section{Background}

In this work, push quantization techniques to their breaking point by scaling transformer models. We are interested in two questions: at which scale and why do quantization techniques fail and how does this related to quantization precision? To answer these questions we study high-precision asymmetric quantization (zeropoint quantization) and symmetric quantization (absolute maximum quantization). While zeropoint quantization offers high precision by using the full bit-range of the datatype, it is rarely used due to practical constraints. Absolute maximum quantization is the most commonly used technique.

\subsection{8-bit Data Types and Quantization}

\textbf{Absmax quantization} scales inputs into the 8-bit range $[-127, 127]$ by multiplying with $s_{x_{f16}}$ which is 127 divided by the absolute maximum of the entire tensor. This is equivalent to dividing by the infinity norm and multiplying by 127. As such, for an FP16 input matrix $\mathbf{X}_{f16}\in \mathbb{R}^{s\times h}$ Int8 absmax quantization is given by:

\[ \mathbf{X}_{i8} = \Biggl{\lfloor}{\dfrac{127\cdot\mathbf{X}_{f16}}{\max\limits_{ij}(|\mathbf{X}_{f16_{ij}}|)}}\Biggr{\rceil}= \Biggl{\lfloor}{\dfrac{127}{\norm{\mathbf{X}_{f16}}_\infty}\mathbf{X}_{f16}}\Biggr{\rceil} = \Bigl{\lfloor}{s_{x_{f16}} \mathbf{X}_{f16}}\Bigr{\rceil},\]
where $\lfloor{} \rceil{}$ indicates rounding to the nearest integer.

\textbf{Zeropoint quantization} shifts the input distribution into the full range $[-127, 127]$ by scaling with the normalized dynamic range $nd_x$ and then shifting by the zeropoint $zp_x$. With this affine transformation, any input tensors will use all bits of the data type, thus {\it reducing the quantization error for asymmetric distributions}. For example, for ReLU outputs, in absmax quantization all values in $[-127, 0)$ go unused, whereas in zeropoint quantization the full $[-127, 127]$ range is used. Zeropoint quantization is given by the following equations:
\begin{equation}
nd_{x_{f16}} = \dfrac{2\cdot 127}{\max\limits_{ij}(\mathbf{X}_{f16}^{ij}) - \min\limits_{ij}(\mathbf{X}_{f16}^{ij})}
\end{equation}
\begin{equation}
zp_{x_{i16}} = \bigl\lfloor{\mathbf{X}_{f16} \cdot \min\limits_{ij}(\mathbf{X}_{f16}^{ij})\bigr\rceil}
\end{equation}
\begin{equation}
\mathbf{X}_{i8} = \bigl\lfloor{{nd_{x_{f16}} \mathbf{X}_{f16}}\bigr\rceil}
\end{equation}
To use zeropoint quantization in an operation we feed both the tensor $\mathbf{X}_{i8}$ and the zeropoint $zp_{x_{i16}}$ into a special instruction\footnote{\url{https://www.felixcloutier.com/x86/pmaddubsw}} which adds $zp_{x_{i16}}$ to each element of $\mathbf{X}_{i8}$ before performing a 16-bit integer operation. For example, to multiply two zeropoint quantized numbers $A_{i8}$ and $B_{i8}$ along with their zeropoints $zp_{a_{i16}}$ and $zp_{b_{i16}}$ we calculate:
\begin{equation}
C_{i32}  = \text{multiply}_{i16}({A}_{zp_{a_{i16}}},{B}_{zp_{b_{i16}}}) = ({A}_{i8}+zp_{a_{i16}})({B}_{i8}+zp_{b_{i16}})
\end{equation}
where unrolling is required if the instruction multiply$_{i16}$ is not available such as on GPUs or TPUs:
\begin{equation}
C_{i32}= {A}_{i8}{B}_{i8} + {A}_{i8}zp_{b_{i16}} + {B}_{i8}zp_{a_{i16}} + zp_{a_{i16}}zp_{b_{i16}},
\end{equation}
where $A_{i8}B_{i8}$ is computed with Int8 precision while the rest is computed in Int16/32 precision. As such, zeropoint quantization can be slow if the multiply$_{i16}$ instruction is not available. In both cases, the outputs are accumulated as a 32-bit integer $C_{i32}$. To dequantize $C_{i32}$, we divide by the scaling constants $nd_{a_{f16}}$ and $nd_{b_{f16}}$.

\paragraph{Int8 Matrix Multiplication with 16-bit Float Inputs and Outputs.}

Given hidden states ${\mathbf{X}_{f16}\in \mathbb{R}^{s\times h}}$ and weights $\mathbf{W}_{f16}\in \mathbb{R}^{h\times o}$ with sequence dimension $s$, feature dimension $h$, and output dimension $o$ we perform 8-bit matrix multiplication with 16-bit inputs and outputs as follows:
\begin{equation}
\begin{split}
    \mathbf{X}_{f16} \mathbf{W}_{f16} =\mathbf{C}_{f16}  & \approx \frac{1}{c_{x_{f16}}c_{w_{f16}}} \mathbf{C}_{i32} = S_{f16}\cdot\mathbf{C}_{i32}\\
    & \approx S_{f16}\cdot\mathbf{A}_{i8}\mathbf{B}_{i8} = S_{f16}\cdot Q(\mathbf{A}_{f16})\phantom{.} Q(\mathbf{B}_{f16}),
\end{split}
\end{equation}
Where $Q(\cdot)$ is either absmax or zeropoint quantization and $c_{x_{f16}}$ and $c_{w_{f16}}$ are the respective tensor-wise scaling constants $s_x$ and $s_w$ for absmax or $nd_x$ and $nd_w$ for zeropoint quantization.

\section{Int8 Matrix Multiplication at Scale}
\label{sec:matmulatscale}

The main challenge with quantization methods that use a single scaling constant per tensor is that a single outlier can reduce the quantization precision of all other values. As such, it is desirable to have multiple scaling constants per tensor, such as block-wise constants \citep{dettmers2022optimizers}, so that the effect of that outliers is confined to each block. We improve upon one of the most common ways of blocking quantization, row-wise quantization \citep{fbgemm}, by using vector-wise quantization, as described in more detail below.

To handle the large magnitude outlier features that occur in all transformer layers beyond the 6.7B scale, vector-wise quantization is no longer sufficient. For this purpose, we develop mixed-precision decomposition, where the small number of large magnitude feature dimensions ($\approx$0.1\%) are represented in 16-bit precision while the other 99.9\% of values are multiplied in 8-bit. Since most entries are still represented in low-precision, we retain about 50\% memory reduction compared to 16-bit. For example, for BLOOM-176B, we reduce the memory footprint of the model by 1.96x.

Vector-wise quantization and mixed-precision decomposition are shown in Figure~\ref{fig:method}. The {\bf LLM.int8()} method is the combination of absmax vector-wise quantization and mixed precision decomposition. 

\subsection{Vector-wise Quantization}

One way to increase the number of scaling constants for matrix multiplication is to view matrix multiplication as a sequence of independent inner products. Given the hidden states $\mathbf{X}_{f16}\in \mathbb{R}^{b\times h}$ and weight matrix $\mathbf{W}_{f16}\in \mathbb{R}^{h\times o}$, we can assign a different scaling constant $c_{x_{f16}}$to each row of $\mathbf{X}_{f16}$ and $c_w$ to each column of $\mathbf{W}_{f16}$. To dequantize, we denormalize each inner product result by $1/(c_{x_{f16}}c_{w_{f16}})$. For the whole matrix multiplication this is equivalent to denormalization by the outer product $\mathbf{c}_{x_{f16}} \otimes \mathbf{c}_{w_{f16}}$, where $\mathbf{c}_x \in \mathbb{R}^s$ and $\mathbf{c}_w \in \mathbb{R}^o$. As such the full equation for matrix multiplication with row and column constants is given by:
\begin{equation}
    \mathbf{C}_{f_{16}} \approx \frac{1}{\mathbf{c}_{x_{f16}}\otimes \mathbf{c}_{w_{f16}}} \mathbf{C}_{i32} = S\cdot\mathbf{C}_{i32}= \mathbf{S}\cdot\mathbf{A}_{i8}\mathbf{B}_{i8} = \mathbf{S}\cdot Q(\mathbf{A}_{f16})\phantom{.} Q(\mathbf{B}_{f16}),
\end{equation}
which we term {\it vector-wise quantization} for matrix multiplication.

\subsection{The Core of LLM.int8(): Mixed-precision Decomposition}

In our analysis, we demonstrate that a significant problem for billion-scale 8-bit transformers is that they have large magnitude features ({\it columns}), which are important for transformer performance and require high precision quantization. However, vector-wise quantization, our best quantization technique, quantizes each {\it row} for the hidden state, which is ineffective for outlier features. Luckily, we see that these outlier features are both incredibly sparse and systematic in practice, making up only about 0.1\% of all feature dimensions, thus allowing us to develop a new decomposition technique that focuses on high precision multiplication for these particular dimensions.

We find that given input matrix $\mathbf{X}_{f16}\in \mathbb{R}^{s\times h}$, these outliers occur systematically for almost all sequence dimensions $s$ but are limited to specific feature/hidden dimensions $h$. As such, we propose {\it mixed-precision decomposition} for matrix multiplication where we separate outlier feature dimensions into the set ${O =\{i| i\in \mathbb{Z}, 0 \leq i \leq h\}}$, which contains all dimensions of $h$ which have at least one outlier with a magnitude larger than the threshold $\alpha$. In our work, we find that $\alpha=6.0$ is sufficient to reduce transformer performance degradation close to zero. Using Einstein notation where all indices are superscripts, given the weight matrix $\mathbf{W}_{f16}\in \mathbb{R}^{h\times o}$, mixed-precision decomposition for matrix multiplication is defined as follows:
\begin{equation}
    \mathbf{C}_{f16} \approx \sum\limits_{h\in O} \mathbf{X}_{f16}^h \mathbf{W}_{f16}^h + \mathbf{S}_{f16} \cdot  \sum\limits_{h\not\in O} \mathbf{X}_{i8}^h \mathbf{W}_{i8}^h
\end{equation}
where $\mathbf{S}_{f16}$ is the denormalization term for the Int8 inputs and weight matrices $\mathbf{X}_{i8}$ and $\mathbf{W}_{i8}$.

This separation into 8-bit and 16-bit allows for high-precision multiplication of outliers while using memory-efficient matrix multiplication with 8-bit weights of more than 99.9\% of values. Since the number of outlier feature dimensions is not larger than 7 ($|O|\leq7$) for transformers up to 13B parameters, this decomposition operation only consumes about 0.1\% additional memory.

\subsection{Experimental Setup}

We measure the robustness of quantization methods as we scale the size of several publicly available pretrained language models up to 175B parameters. The key question is not how well a quantization method performs for a particular model but the trend of how such a method performs as we scale.

We use two setups for our experiments. One is based on language modeling perplexity, which we find to be a highly robust measure that is very sensitive to quantization degradation. We use this setup to compare different quantization baselines. Additionally, we evaluate zeroshot accuracy degradation on OPT models for a range of different end tasks, where we compare our methods with a 16-bit baseline.

For the language modeling setup, we use dense autoregressive transformers pretrained in fairseq \citep{ott2019fairseq} ranging between 125M and 13B parameters. These transformers have been pretrained on Books~\citep{zhu2015aligning}, English Wikipedia, CC-News \citep{nagel2016cc}, OpenWebText \citep{gokaslan2019openwebtext}, CC-Stories \citep{trinh2018simple}, and English CC100 \citep{wenzek-etal-2020-ccnet}. For more information on how these pretrained models are trained, see \citet{artetxe2021efficient}.
 
To evaluate the language modeling degradation after Int8 quantization, we evaluate the perplexity of the 8-bit transformer on validation data of the C4 corpus \citep{raffel2019exploring} which is a subset of the Common Crawl corpus.\footnote{\url{https://commoncrawl.org/}} We use NVIDIA A40 GPUs for this evaluation.

To measure degradation in zeroshot performance, we use OPT models \citep{zhang2022opt}, and we evaluate these models on the EleutherAI language model evaluation harness \citep{eval-harness}.

\subsection{Main Results}

The main language modeling perplexity results on the 125M to 13B Int8 models evaluated on the C4 corpus can be seen in Table~\ref{tbl:inference}. We see that absmax, row-wise, and zeropoint quantization fail as we scale, where models after 2.7B parameters perform worse than smaller models. Zeropoint quantization fails instead beyond 6.7B parameters. Our method, LLM.int8(), is the only method that preserves perplexity. As such, LLM.int8() is the only method with a favorable scaling trend.

When we look at the scaling trends of zeroshot performance of OPT models on the EleutherAI language model evaluation harness in Figure~\ref{fig:results}, we see that LLM.int8() maintains full 16-bit performance as we scale from 125M to 175B parameters. On the other hand, the baseline, 8-bit absmax vector-wise quantization, scales poorly and degenerates into random performance.

\begin{table}[]
\centering
\caption{C4 validation perplexities of quantization methods for different transformer sizes ranging from 125M to 13B parameters. We see that absmax, row-wise, zeropoint, and vector-wise quantization leads to significant performance degradation as we scale, particularly at the 13B mark where 8-bit 13B perplexity is worse than 8-bit 6.7B perplexity. If we use LLM.int8(), we recover full perplexity as we scale. Zeropoint quantization shows an advantage due to asymmetric quantization but is no longer advantageous when used with mixed-precision decomposition.}
\label{tbl:inference}
\begin{tabular}{lccccc}\toprule
Parameters &
  \multicolumn{1}{l}{125M} &
  \multicolumn{1}{l}{1.3B} &
  \multicolumn{1}{l}{2.7B} &
  \multicolumn{1}{l}{6.7B} &
  \multicolumn{1}{l}{13B} \\
 \toprule
32-bit Float                 & 25.65 & 15.91 & 14.43 & 13.30  & 12.45 \\\midrule
Int8 absmax                   & 87.76 & 16.55 & 15.11 & 14.59 & 19.08 \\
Int8 zeropoint                & 56.66 & 16.24 & 14.76 & 13.49 & 13.94 \\\midrule
Int8 absmax row-wise                   & 30.93 & 17.08 & 15.24 & 14.13 & 16.49 \\
Int8 absmax vector-wise                   & 35.84 & 16.82 & 14.98 & 14.13 & 16.48 \\
Int8 zeropoint vector-wise         & 25.72 & 15.94 & 14.36 & 13.38 & 13.47 \\\midrule
Int8 absmax row-wise + decomposition                     & 30.76 & 16.19 & 14.65 & 13.25 & 12.46 \\
Absmax LLM.int8() (vector-wise + decomp)          & 25.83 & 15.93 & 14.44 & \bf 13.24 & \bf 12.45 \\
Zeropoint LLM.int8() (vector-wise + decomp)          & \bf 25.69 & \bf 15.92 & \bf 14.43 & \bf 13.24 & \bf 12.45\\\bottomrule
\end{tabular}
\end{table}

Although our primary focus is on saving memory, we also measured the run time of LLM.int8(). The quantization overhead can slow inference for models with less than 6.7B parameters, as compared to a FP16 baseline. However, models of 6.7B parameters or less fit on most GPUs and quantization is less needed in practice. LLM.int8() run times is about two times faster for large matrix multiplications equivalent to those in 175B models. Appendix~\ref{app:inference} provides more details on these experiments. 

\section{Emergent Large Magnitude Features in Transformers at Scale}
\label{sec:emergence}

As we scale transformers, outlier features with large magnitudes emerge and strongly affect {\it all} layers and their quantization. Given a hidden state $\mathbf{X}\in \mathbb{R}^{s\times h}$ where $s$ is the sequence/token dimension and $h$ the hidden/feature dimension, we define a feature to be a particular dimension $h_i$. Our analysis looks at a particular feature dimension $h_i$ across all layers of a given transformer.

We find that outlier features strongly affect attention and the overall predictive performance of transformers. While up to 150k outliers exist per 2048 token sequence for a 13B model, these outlier features are highly systematic and only representing at most 7 unique feature dimensions $h_i$. Insights from this analysis were critical to developing mixed-precision decomposition. Our analysis explains the advantages of zeropoint quantization and why they disappear with the use of mixed-precision decomposition and the quantization performance of small vs. large models.

\subsection{Finding Outlier Features}
\vspace{-0.5em}

The difficulty with the quantitative analysis of emergent phenomena is two-fold. We aim to select a small subset of features for analysis such that the results are intelligible and not to complex while also capturing important probabilistic and structured patterns. We use an empirical approach to find these constraints.
We define outliers according to the following criteria: the magnitude of the feature is at least 6.0, affects at least 25\% of layers, and affects at least 6\% of the sequence dimensions. 

More formally, given a transformer with $L$ layers and hidden state $\mathbf{X}_l\in \mathbb{R}^{s\times h}, l=0...L$ where $s$ is the sequence dimension and $h$ the feature dimension, we define a feature to be a particular dimension $h_i$ in any of the hidden states $\mathbf{X}_{l_{i}}$. We track dimensions $h_i, 0 \leq i \leq h$, which have at least one value with a magnitude of $\alpha\geq6$ and we only collect statistics if these outliers occur in the {\it same} feature dimension $h_i$ in at least 25\% of transformer layers $0...L$ and appear in at least 6\% of all sequence dimensions $s$ across all hidden states $\mathbf{X}_l$. Since feature outliers only occur in attention projection (key/query/value/output) and the feedforward network expansion layer (first sub-layer), we ignore the attention function and the FFN contraction layer (second sub-layer) for this analysis.

Our reasoning for these thresholds is as follows. We find that using mixed-precision decomposition, perplexity degradation stops if we treat any feature with a magnitude 6 or larger as an outlier feature. 
For the number of layers affected by outliers, we find that outlier features are {\it systematic} in large models: they either occur in most layers or not at all. On the other hand, they are {\it probabilistic} in small models: they occur {\it sometimes} in {\it some} layers for each sequence. 
As such, we set our threshold for how many layers need to be affected to detect an outlier feature in such a way as to limit detection to a {\it single} outlier in our smallest model with 125M parameters. This threshold corresponds to that at least 25\% of transformer layers are affected by an outlier in the same feature dimension. The second most common outlier occurs in only a single layer (~2\% of layers), indicating that this is a reasonable threshold.
We use the same procedure to find the threshold for how many sequence dimensions are affected by outlier features in our 125M model: outliers occur in at least 6\% of sequence dimensions.

We test models up to a scale of 13B parameters. To make sure that the observed phenomena are not due to bugs in software, we evaluate transformers that were trained in three different software frameworks. We evaluate four GPT-2 models which use OpenAI software, five Meta AI models that use Fairseq \citep{ott2019fairseq}, and one EleutherAI model GPT-J that uses Tensorflow-Mesh \citep{shazeer2018mesh}. More details can be found in Appendix~\ref{appendix:outliers}. We also perform our analysis in two different inference software frameworks: Fairseq and Hugging Face Transformers \citep{wolf2019huggingface}. 

\subsection{Measuring the Effect of Outlier Features}

To demonstrate that the outlier features are essential for attention and predictive performance, we set the outlier features to zero before feeding the hidden states $\mathbf{X}_l$ into the attention projection layers and then compare the top-1 softmax probability with the regular softmax probability with outliers. We do this for all layers independently, meaning we forward the regular softmax probabilities values to avoid cascading errors and isolate the effects due to the outlier features.
We also report the perplexity degradation if we remove the outlier feature dimension (setting them to zero) and propagate these altered, hidden states through the transformer. As a control, we apply the same procedure for random non-outlier feature dimensions and note attention and perplexity degradation.

\begin{figure}
\centering
  \begin{subfigure}[b]{0.45\textwidth}
         \includegraphics[scale=0.55]{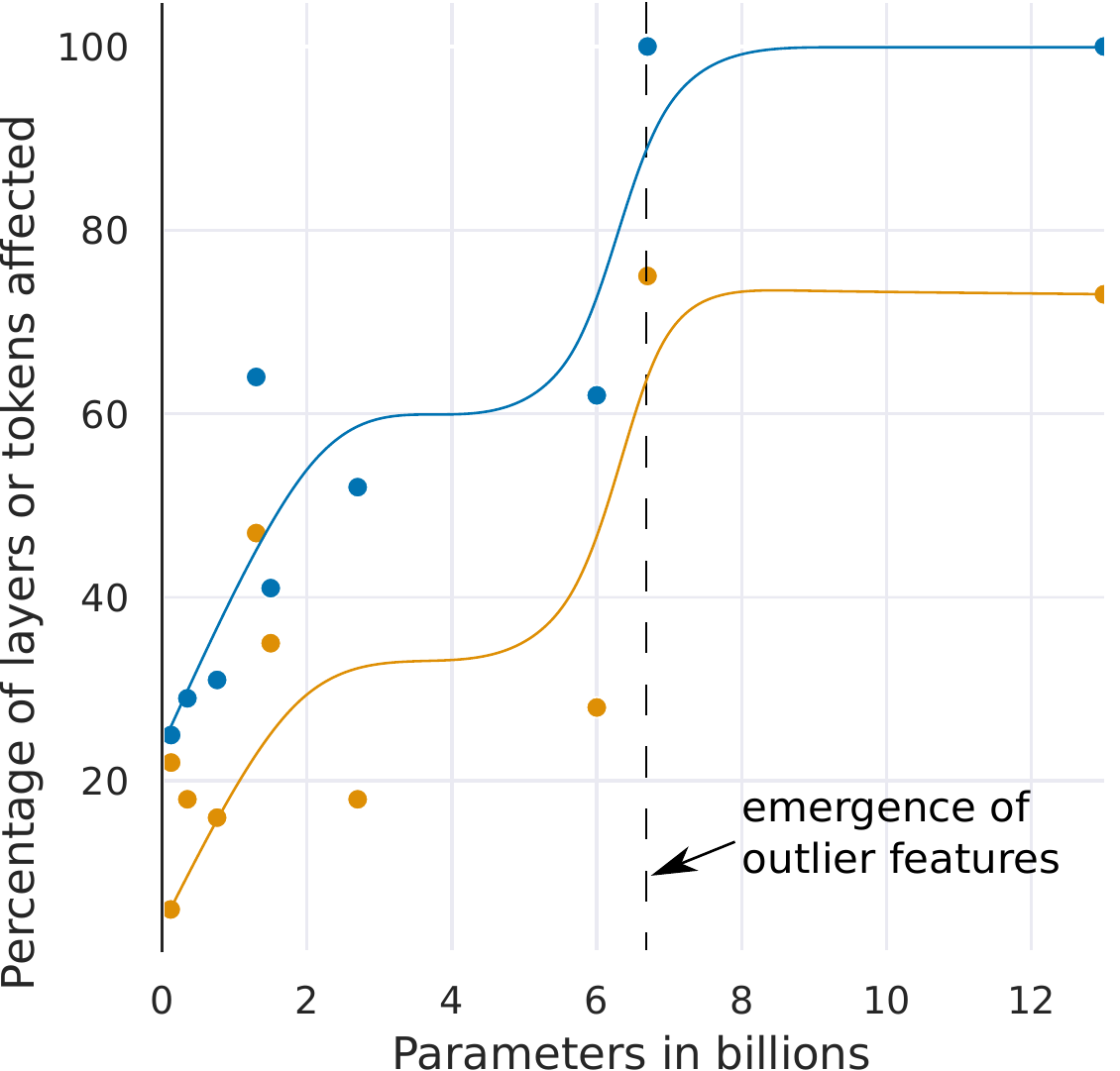}
        \caption{}
        \label{fig:outlier_param}
     \end{subfigure}
     \hfill
     \begin{subfigure}[b]{0.45\textwidth}
     \includegraphics[scale=0.55]{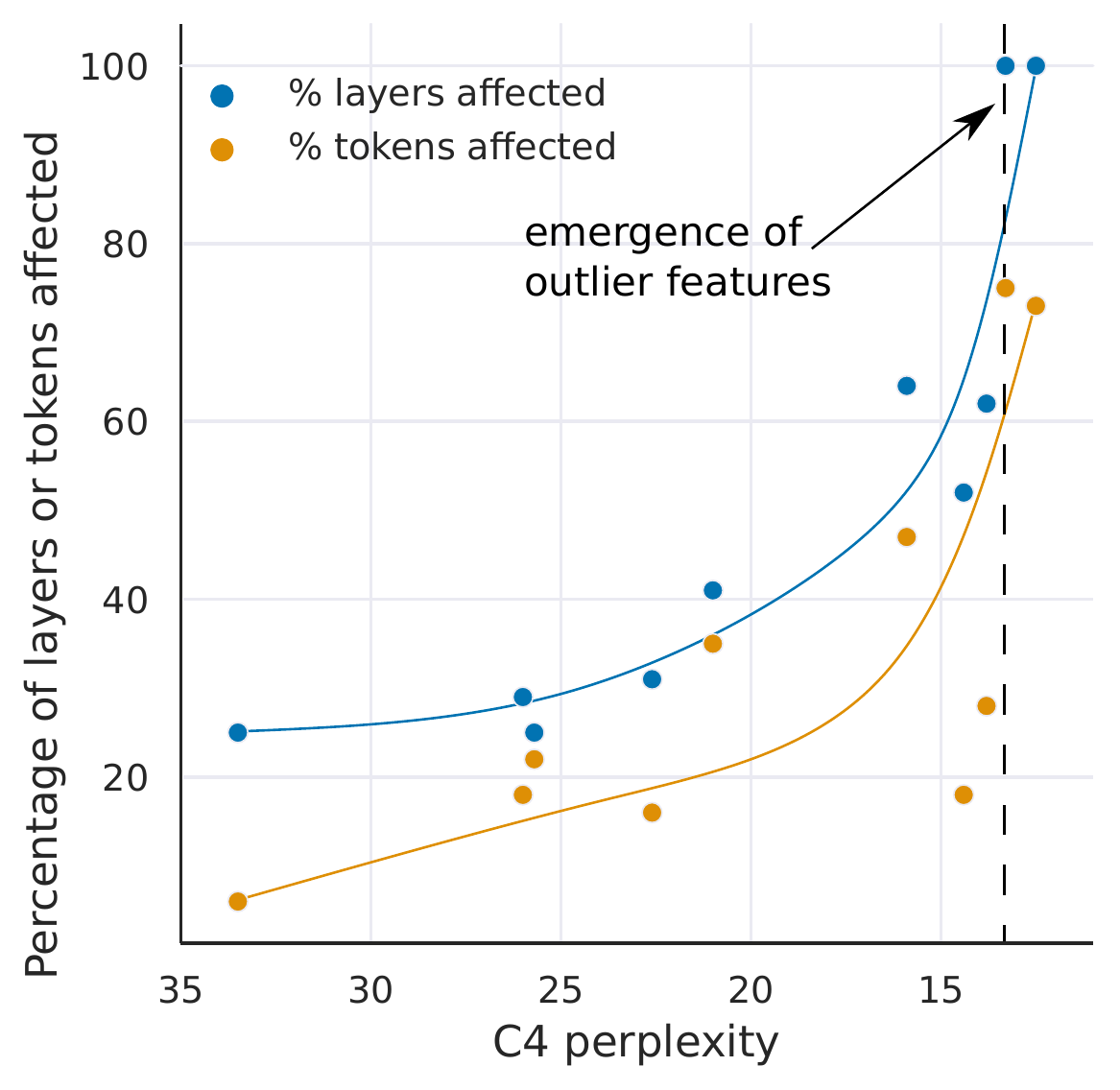}
        \caption{}
        \label{fig:outlier_ppl}
     \end{subfigure}
        \caption{Percentage of layers and all sequence dimensions affected by large magnitude outlier features across the transformer by (a) model size or (b) C4 perplexity. Lines are B-spline interpolations of 4 and 9 linear segments for (a) and (b). Once the phase shift occurs, outliers are present in all layers and in about 75\% of all sequence dimensions. While (a) suggest a sudden phase shift in parameter size, (b) suggests a gradual exponential phase shift as perplexity decreases. The stark shift in (a) co-occurs with the sudden degradation of performance in quantization methods.}
        \label{fig:emergence}
\end{figure}

\begin{figure}
\centering
  \begin{subfigure}[b]{0.45\textwidth}
         \includegraphics[scale=0.46]{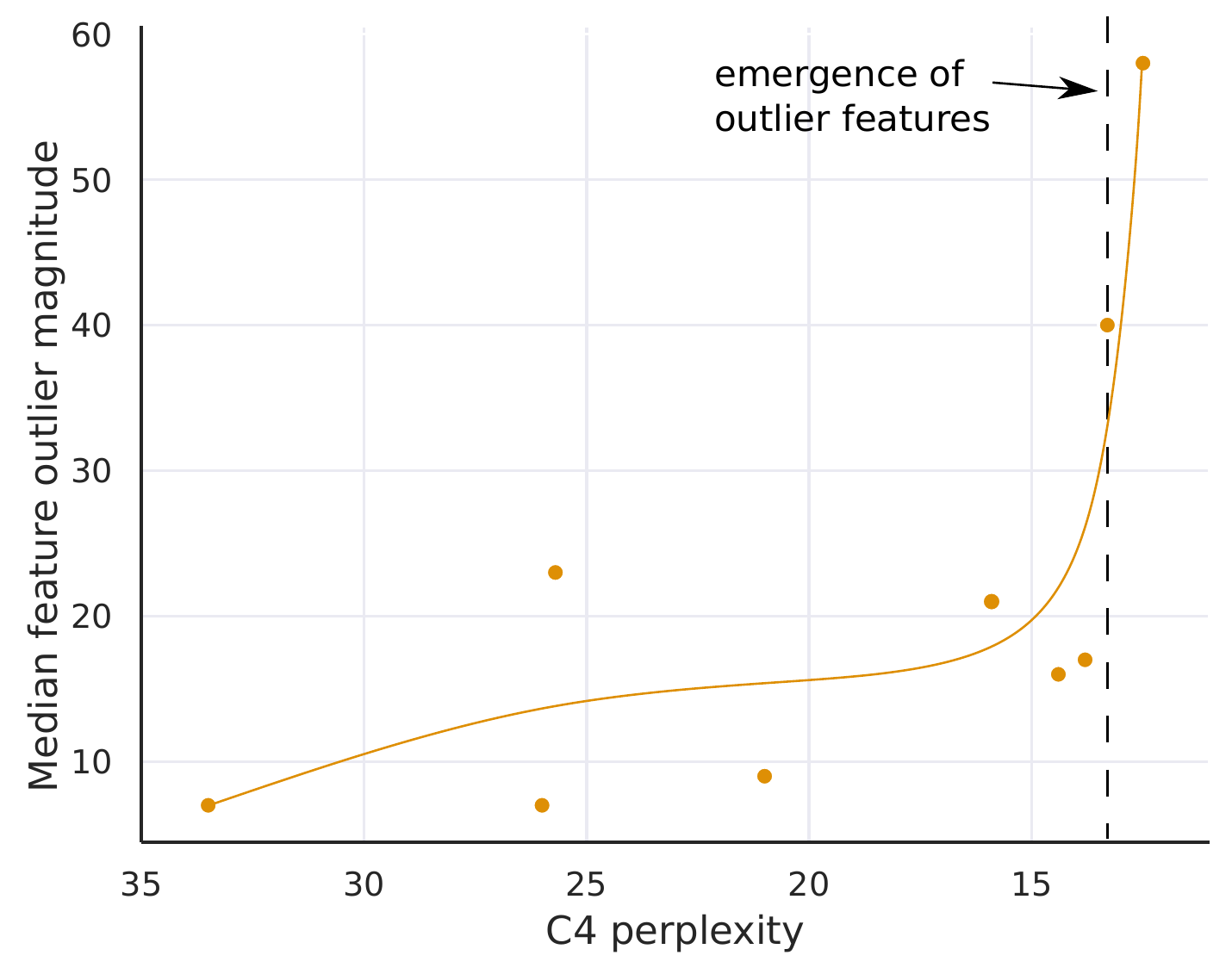}
        \caption{}
        \label{fig:quartiles}
     \end{subfigure}
     \hfill
     \begin{subfigure}[b]{0.45\textwidth}
     \includegraphics[scale=0.45]{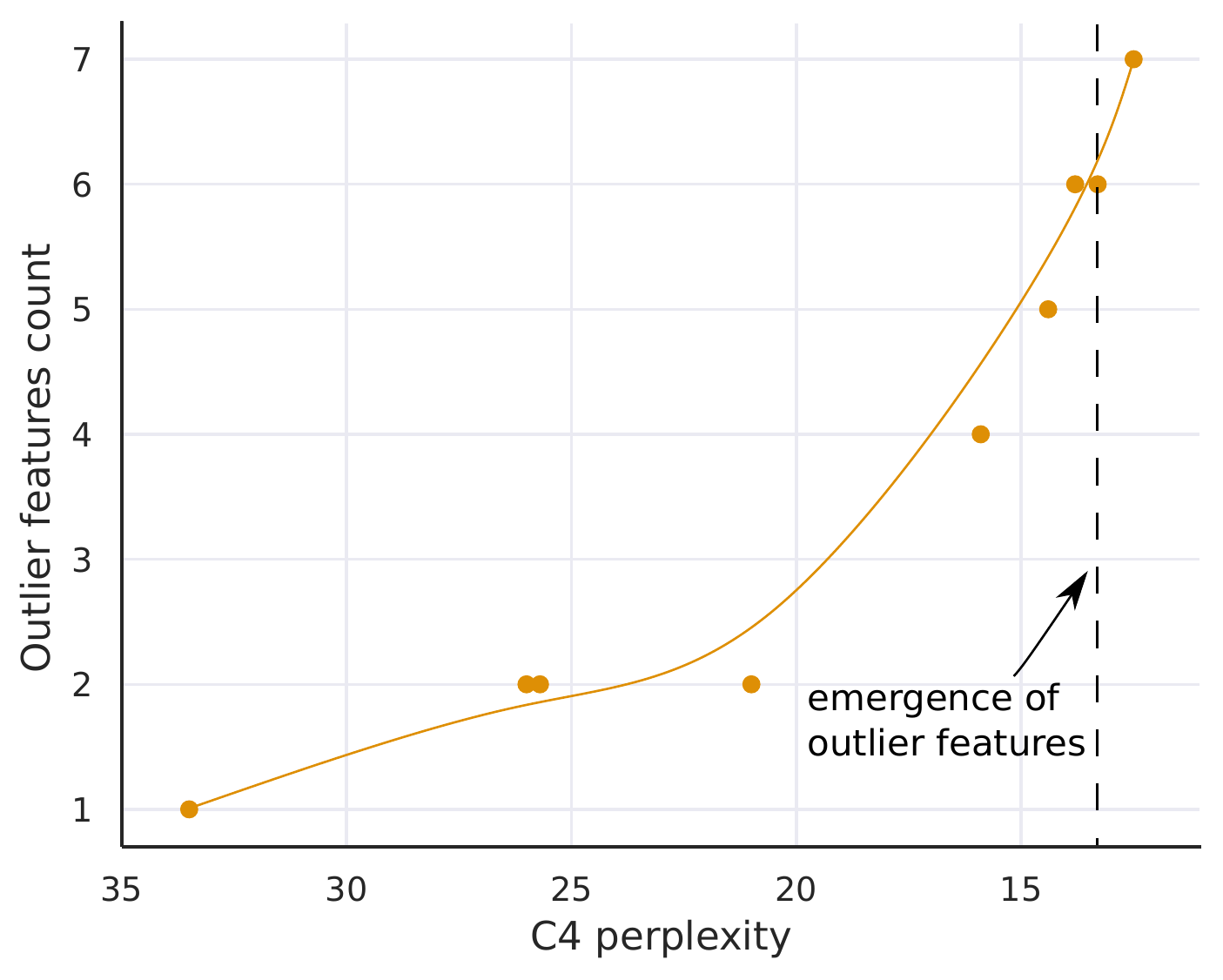}
        \caption{}
        \label{fig:outlier_count}
     \end{subfigure}
        \caption{The median magnitude of the largest outlier feature in (a) indicates a sudden shift in outlier size. This appears to be the prime reason why quantization methods fail after emergence. While the number of outlier feature dimensions is only roughly proportional to model size, (b) shows that the number of outliers is {\it strictly monotonic} with respect to perplexity across all models analyzed. Lines are B-spline interpolations of 9 linear segments.}
        \label{fig:outliers}
\end{figure}

Our main quantitative results can be summarized as four main points. 

(1) When measured by the number of parameters, the emergence of large magnitude features across {\it all} layers of a transformer occurs suddenly between 6B and 6.7B parameters as shown in Figure~\ref{fig:outlier_param} as the percentage of layers affected increases from 65\% to 100\%. The number of sequence dimensions affected increases rapidly from 35\% to 75\%. This sudden shift co-occurs with the point where quantization begins to fail.

(2) Alternatively, when measured by perplexity, the emergence of large magnitude features across all layers of the transformer can be seen as emerging smoothly according to an exponential function of decreasing perplexity, as seen in Figure~\ref{fig:outlier_ppl}. This indicates that there is nothing sudden about emergence and that we might be able to detect emergent features before a phase shift occurs by studying exponential trends in smaller models. This also suggests that emergence is not only about model size but about perplexity, which is related to multiple additional factors such as the amount of training data used, and data quality \citep{hoffmann2022training,henighan2020scaling}.

(3) Median outlier feature magnitude rapidly increases once outlier features occur in all layers of the transformer, as shown in Figure~\ref{fig:quartiles}. The large magnitude of outliers features and their asymmetric distribution disrupts Int8 quantization precision. This is the core reason why quantization methods fail starting at the 6.7B scale -- the range of the quantization distribution is too large so that most quantization bins are empty and small quantization values are quantized to zero, essentially extinguishing information.  We hypothesize that besides Int8 inference, regular 16-bit floating point training becomes unstable due to outliers beyond the 6.7B scale -- it is easy to exceed the maximum 16-bit value 65535 by chance if you multiply by vectors filled with values of magnitude 60.

(4) The number of outliers features increases strictly monotonically with respect to decreasing C4 perplexity as shown in Figure~\ref{fig:outlier_count}, while a relationship with model size is non-monotonic. This indicates that model perplexity rather than mere model size determines the phase shift. We hypothesize that model size is only one important covariate among many that are required to reach emergence.

These outliers features are highly systematic after the phase shift occurred. For example, for a 6.7B transformer with a sequence length of 2048, we find about 150k outlier features per sequence for the entire transformer, but these features are concentrated in only 6 different hidden dimensions. 

These outliers are critical for transformer performance. If the outliers are removed, the mean top-1 softmax probability is reduced from about 40\% to about 20\%, and validation perplexity increases by 600-1000\% even though there are at most 7 outlier feature dimensions. When we remove 7 random feature dimensions instead, the top-1 probability decreases only between 0.02-0.3\%, and perplexity increases by 0.1\%. This highlights the critical nature of these feature dimensions. Quantization precision for these outlier features is paramount as even tiny errors greatly impact model performance.

\subsection{Interpretation of Quantization Performance}

Our analysis shows that outliers in particular feature dimensions are ubiquitous in large transformers, and these feature dimensions are critical for transformer performance.
Since row-wise and vector-wise quantization scale each hidden state sequence dimension $s$ (rows) and because outliers occur in the feature dimension $h$ (columns), both methods cannot deal with these outliers effectively. This is why absmax quantization methods fail quickly after emergence.

However, almost all outliers have a strict asymmetric distribution: they are either solely positive or negative (see Appendix~\ref{appendix:outliers}). This makes zeropoint quantization particularly effective for these outliers, as zeropoint quantization is an asymmetric quantization method that scales these outliers into the full $[-127, 127]$ range. This explains the strong performance in our quantization scaling benchmark in Table~\ref{tbl:inference}. However, at the 13B scale, even zeropoint quantization fails due to accumulated quantization errors and the quick growth of outlier magnitudes, as seen in Figure~\ref{fig:quartiles}.

If we use our full LLM.int8() method with mixed-precision decomposition, the advantage of zeropoint quantization disappears indicating that the remaining decomposed features are symmetric. However, vector-wise still has an advantage over row-wise quantization, indicating that the enhanced quantization precision of the model weights is needed to retain full precision predictive performance.

\section{Related work}
\vspace{-1em}

There is closely related work on quantization data types and quantization of transformers, as described below. Appendix~\ref{appendix:relatedwork} provides further related work on quantization of convolutional networks.

\textbf{8-bit Data Types.} Our work studies quantization techniques surrounding the Int8 data type, since it is currently the only 8-bit data type supported by GPUs. Other common data types are fixed point or floating point 8-bit data types (FP8). These data types usually have a sign bit and different exponent and fraction bit combinations. For example, a common variant of this data type has 5 bits for the exponent and 2 bits for the fraction \citep{wang2018training8bit,sun2019hybrid8bit,cambier2020shiftsqueeze,mellempudi2019bit8} and uses either no scaling constants or zeropoint scaling. These data types have large errors for large magnitude values since they have only 2 bits for the fraction but provide high accuracy for small magnitude values.
\citet{jin2022f8net} provide an excellent analysis of when certain fixed point exponent/fraction bit widths are optimal for inputs with a particular standard deviation. We believe FP8 data types offer superior performance compared to the Int8 data type, but currently, neither GPUs nor TPUs support this data type.

\textbf{Outlier Features in Language Models.} Large magnitude outlier features in language models have been studied before \citep{timkey2021all,bondarenko2021understanding, wei2022outlier,luo-etal-2021-positional}. Previous work proved the theoretical relationship between outlier appearance in transformers and how it relates to layer normalization and the token frequency distribution \citep{gao2019representation}. Similarly, \citet{kovaleva2021bert} attribute the appearance of outliers in BERT model family to LayerNorm, and \citet{puccetti2022outliers} show empirically that outlier emergence is related to the frequency of tokens in the training distribution. We extend this work further by showing how the scale of autoregressive models relates to the emergent properties of these outlier features, and showing how appropriately modeling outliers is critical to effective quantization. 

\textbf{Multi-billion Scale Transformer Quantization.} There are two methods that were developed in parallel to ours: nuQmm \citep{park2022nuqmm} and ZeroQuant \citep{yao2022zeroquant}. Both use the same quantization scheme: group-w2ise quantization, which has even finer quantization normalization constant granularity than vector-wise quantization. This scheme offers higher quantization precision but also requires custom CUDA kernels. Both nuQmm and ZeroQuant aim to accelerate inference and reduce the memory footprint while we focus on preserving predictive performance under an 8-bit memory footprint. The largest models that nuQmm and ZeroQuant evaluate are 2.7B and 20B parameter transformers, respectively. ZeroQuant achieves zero-degradation performance for 8-bit quantization of a 20B model. We show that our method allows for zero-degradation quantization of models up to 176B parameters. Both nuQmm and ZeroQuant suggest that finer quantization granularity can be an effective means to quantize large models. These methods are complementary with LLM.int8(). Another parallel work is GLM-130B which uses insights from our work to achieve zero-degradation 8-bit quantization \citep{zeng2022glm}. GLM-130B performs full 16-bit precision matrix multiplication with 8-bit weight storage.

\section{Discussion and Limitations}
\vspace{-1em}
We have demonstrated for the first time that multi-billion parameter transformers can be quantized to Int8 and used immediately for inference without performance degradation. We achieve this by using our insights from analyzing emergent large magnitude features at scale to develop mixed-precision decomposition to isolate outlier features in a separate 16-bit matrix multiplication. In conjunction with vector-wise quantization that yields our method, LLM.int8(), which we show empirically can recover the full inference performance of models with up to  175B parameters.

The main limitation of our work is that our analysis is solely on the Int8 data type, and we do not study 8-bit floating-point (FP8) data types. Since current GPUs and TPUs do not support this data type, we believe this is best left for future work. However, we also believe many insights from Int8 data types will directly translate to FP8 data types.
Another limitation is that we only study models with up to 175B parameters. While we quantize a 175B model to Int8 without performance degradation, additional emergent properties might disrupt our quantization methods at larger scales.

A third limitation is that we do not use Int8 multiplication for the attention function. Since our focus is on reducing the memory footprint and the attention function does not use any parameters, it was not strictly needed. However, an initial exploration of this problem indicated that a solution required additional quantization methods beyond those we developed here, and we leave this for future work.

A final limitation is that we focus on inference but do not study training or finetuning. We provide an initial analysis of Int8 finetuning and training at scale in Appendix~\ref{appendix:training}. Int8 training at scale requires complex trade-offs between quantization precision, training speed, and engineering complexity and represents a very difficult problem. We again leave this to future work. 

\begin{table}[h]
\centering
\caption{Different hardware setups and which methods can be run in 16-bit vs. 8-bit precision. We can see that our 8-bit method makes many models accessible that were not accessible before, in particular, OPT-175B/BLOOM.}
\label{tbl:memsavings}
\resizebox{\textwidth}{!}{%
\begin{tabular}{lllcc}\toprule
                 &             &            & \multicolumn{2}{c}{Largest Model that can be run}     \\\cmidrule{4-5}
Class            & Hardware    & GPU Memory & 8-bit                     & 16-bit                    \\\midrule
Enterprise       & 8x A100     & 80 GB      & \textbf{OPT-175B / BLOOM} & \textbf{OPT-175B / BLOOM} \\
Enterprise       & 8x A100     & 40 GB      & \textbf{OPT-175B / BLOOM} & OPT-66B                   \\
Academic server  & 8x RTX 3090 & 24 GB      & \textbf{OPT-175B / BLOOM} & OPT-66B                   \\
Academic desktop & 4x RTX 3090 & 24 GB      & \textbf{OPT-66B}          & OPT-30B                   \\
Paid Cloud       & Colab Pro   & 15 GB      & \textbf{OPT-13B}          & GPT-J-6B                  \\
Free Cloud       & Colab       & 12 GB      & \textbf{T0/T5-11B}        & GPT-2 1.3B  \\\bottomrule             
\end{tabular}%
}
\end{table}

\section{Broader Impacts}
\vspace{-1em}

The main impact of our work is enabling access to large models that previously could not fit into GPU memory. This enables research and applications which were not possible before due to limited GPU memory, in particular for researchers with the least resources. See Table~\ref{tbl:memsavings} for model/GPU combinations which are now accessible without performance degradation.
However, our work also enables resource-rich organizations with many GPUs to serve more models on the same number of GPUs, which might increase the disparities between resource-rich and poor organizations.

In particular, we believe that the public release of large pretrained models, for example, the recent Open Pretrained Transformers (OPT) \citep{zhang2022opt}, along with our new Int8 inference for zero- and few-shot prompting, will enable new research for academic institutions that was not possible before due to resource constraints. The widespread accessibility of such large-scale models will likely have both beneficial and detrimental effects on society that are difficult to predict.

\paragraph{Acknowledgments}

We thank Ofir Press, Gabriel Ilharco, Daniel Jiang, Mitchell Wortsman, Ari Holtzman, Mitchell Gordon for their feedback on drafts of this work. We thank JustHeuristic (Yozh) and Titus von Köller for help with Hugging Face Transformers integration.

\bibliography{references}

\section*{Checklist}

The checklist follows the references.  Please
read the checklist guidelines carefully for information on how to answer these
questions.  For each question, change the default \answerTODO{} to \answerYes{},
\answerNo{}, or \answerNA{}.  You are strongly encouraged to include a {\bf
justification to your answer}, either by referencing the appropriate section of
your paper or providing a brief inline description.  For example:
\begin{itemize}
  \item Did you include the license to the code and datasets? \answerYes{See Section~\ref{gen_inst}.}
  \item Did you include the license to the code and datasets? \answerNo{The code and the data are proprietary.}
  \item Did you include the license to the code and datasets? \answerNA{}
\end{itemize}
Please do not modify the questions and only use the provided macros for your
answers.  Note that the Checklist section does not count towards the page
limit.  In your paper, please delete this instructions block and only keep the
Checklist section heading above along with the questions/answers below.

\begin{enumerate}

\item For all authors...
\begin{enumerate}
  \item Do the main claims made in the abstract and introduction accurately reflect the paper's contributions and scope? \answerYes{}
  \item Did you describe the limitations of your work? \answerYes{See the limitation section}
  \item Did you discuss any potential negative societal impacts of your work?\answerYes{See the Broader Impacts section}
  \item Have you read the ethics review guidelines and ensured that your paper conforms to them?\answerYes{Yes, we believe our work conforms to these guidelines.}
\end{enumerate}

\item If you are including theoretical results...
\begin{enumerate}
  \item Did you state the full set of assumptions of all theoretical results?
    \answerNA{}
        \item Did you include complete proofs of all theoretical results?
    \answerNA{}
\end{enumerate}

\item If you ran experiments...
\begin{enumerate}
  \item Did you include the code, data, and instructions needed to reproduce the main experimental results (either in the supplemental material or as a URL)?
    \answerYes{We will include our code in the supplemental material.}
  \item Did you specify all the training details (e.g., data splits, hyperparameters, how they were chosen)?\answerYes{See the experimental setup section}
        \item Did you report error bars (e.g., with respect to the random seed after running experiments multiple times)?
    \answerNo{Our experiments are deterministic for each model. Instead of running the same model multiple times, we run multiple models at different scales. We are unable to compute error bars for these experiments.}
        \item Did you include the total amount of compute and the type of resources used (e.g., type of GPUs, internal cluster, or cloud provider)?
    \answerYes{See the exper2imental setup section}
\end{enumerate}

\item If you are using existing assets (e.g., code, data, models) or curating/releasing new assets...
\begin{enumerate}
  \item If your work uses existing assets, did you cite the creators?
    \answerYes{See experimental setup section}
  \item Did you mention the license of the assets?
    \answerNo{The license is permissible for all the assets that we use. The individual licenses can easily be looked up.}
  \item Did you include any new assets either in the supplemental material or as a URL?
    \answerNA{}{We only use existing datasets.}
  \item Did you discuss whether and how consent was obtained from people whose data you're using/curating?
    \answerNA{}{}
  \item Did you discuss whether the data you are using/curating contains personally identifiable information or offensive content?
    \answerNA{}{}
\end{enumerate}

\item If you used crowdsourcing or conducted research with human subjects...
\begin{enumerate}
  \item Did you include the full text of instructions given to participants and screenshots, if applicable?
    \answerNA{}{}
  \item Did you describe any potential participant risks, with links to Institutional Review Board (IRB) approvals, if applicable?
    \answerNA{}{}
  \item Did you include the estimated hourly wage paid to participants and the total amount spent on participant compensation?
    \answerNA{}{}
\end{enumerate}

\end{enumerate}

\newpage
\appendix

\section{Memory usage compared to 16-bit precision}

Table~\ref{tbl:memsavings} compares the memory footprint of 16-bit inference and LLM.int8() for different open source models. We can see, that LLM.int8() allows to run the largest open source models OPT-175B and BLOOM-176B on a single node equipped with consumer-grade GPUs.

\begin{table}[h]
\centering
\caption{Different hardware setups and which methods can be run in 16-bit vs. 8-bit precision. We can see that our 8-bit method makes many models accessible that were not accessible before, in particular, OPT-175B/BLOOM.}
\label{tbl:memsavings}
\resizebox{\textwidth}{!}{%
\begin{tabular}{lllcc}\toprule
                 &             &            & \multicolumn{2}{c}{Largest Model that can be run}     \\\cmidrule{4-5}
Class            & Hardware    & GPU Memory & 8-bit                     & 16-bit                    \\\midrule
Enterprise       & 8x A100     & 80 GB      & \textbf{OPT-175B / BLOOM} & \textbf{OPT-175B / BLOOM} \\
Enterprise       & 8x A100     & 40 GB      & \textbf{OPT-175B / BLOOM} & OPT-66B                   \\
Academic server  & 8x RTX 3090 & 24 GB      & \textbf{OPT-175B / BLOOM} & OPT-66B                   \\
Academic desktop & 4x RTX 3090 & 24 GB      & \textbf{OPT-66B}          & OPT-30B                   \\
Paid Cloud       & Colab Pro   & 15 GB      & \textbf{OPT-13B}          & GPT-J-6B                  \\
Free Cloud       & Colab       & 12 GB      & \textbf{T0/T5-11B}        & GPT-2 1.3B  \\\bottomrule             
\end{tabular}%
}
\end{table}

\section{Additional Related Work}
\label{appendix:relatedwork}

\paragraph{Quantization of Transformers with fewer than 1B Parameters}

Quantization of transformers has been focused on sub-billion parameter masked language  model (MLMs), including  BERT~\citep{devlin2018bert} and RoBERTa~\citep{liu2019roberta}.
Versions of 8-bit BERT/RoBERTa include Q8BERT~\citep{zafrir2019q8bert}, QBERT~\citep{shen2020q}, product quantization with quantization noise~\citep{fan2020training}, TernaryBERT~\citep{Zhang2020TernaryBERTDU}, and BinaryBERT~\citep{Bai2021BinaryBERTPT}. Work by \citet{zhao2021automatic} performs both quantization and pruning. All these models require either quantization-aware finetuning or post-training quantization to make the model usable in low-precision. 
In contrast with our methods, the model can be used directly without performance degradation.

If one views matrix multiplication as 1x1 convolution, vector-wise quantization is equivalent to channel-wise quantization for convolution combined with row quantization \citep{fbgemm}. For matrix multiplication, this was used by \citet{wu2020integer} for BERT-sized transformers (350M parameters), while we are the first to study vector-wise quantization for autoregressive and large-scale models.
The only other work that we are aware of that quantizes transformers other than BERT is \citet{chen2020statistical}, which uses post-training quantization with zeropoint quantization in the forward pass and zeropoint-row-wise quantization in the backward pass. However, this work is still for sub-billion parameter transformers. We compare with both zeropoint and row-wise quantization in our evaluations and do not require post-training quantization.

\paragraph{Low-bitwidth and Convolutional Network Quantization}

Work that uses less than 8-bits for data types is usually for convolutional networks (CNNs) to reduce their memory footprint and increase inference speed for mobile devices while minimizing model degradation. Methods for different bit-widths have been studied: 1-bit methods \citep{courbariaux2016bit1,Rastegari2016xnor,courbariaux2015binaryconnect}, 2 to 3-bit \citep{zhu2017ternary,choi2019bit2}, 4-bits \citep{li2019bit4}, more bits \citep{courbariaux2014training}, or a variable amount of bits \citep{gong2019softquant}. For additional related work, please see the survey of \citet{qin2020survey1bit}. While we believe that lower than 8-bit width with some performance degradation is possible for billion-scale transformers, we focus on 8-bit transformers that {\it do not} degrade performance and that can benefit from commonly used GPUs that accelerates inference through Int8 tensor cores.

Another line of work that focuses on convolutional network quantization is to learn adjustments to the quantization procedure to improve quantization errors. For example, using Hessian information \citep{dong2019hawq}, step-size quantization \citep{esser2019learned}, soft quantization \citep{gong2019softquant}, mixed-precision via linear programming optimization \citep{yao2021hawq}, and other learned quantization methods \citep{zhang2018lq, gholami2021survey}.

\section{Detailed Outlier Feature Data}
\label{appendix:outliers}

Table~\ref{tbl:outliers} provides tabulated data from our outlier feature analysis. We provide the quartiles of the most common outlier in each transformer and the number of outliers that are one-sided, that is, which have asymmetric distributions which do not cross zero.

\begin{table}[]
\centering
\caption{Summary statistics of outliers with a magnitude of at least 6 that occur in at least 25\% of all layers and at least 6\% of all sequence dimensions. We can see that the lower the C4 validation perplexity, the more outliers are present. Outliers are usually one-sided, and their quartiles with maximum range show that the outlier magnitude is 3-20x larger than the largest magnitude of other feature dimensions, which usually have a range of [-3.5, 3.5]. With increasing scale, outliers become more and more common in all layers of the transformer, and they occur in almost all sequence dimensions. A phase transition occurs at 6.7B parameters when the same outlier occurs in all layers in the same feature dimension for about 75\% of all sequence dimensions (SDim). Despite only making up about 0.1\% of all features, the outliers are essential for large softmax probabilities. The mean top-1 softmax probability shrinks by about 20\% if outliers are removed. Because the outliers have mostly asymmetric distributions across the sequence dimension $s$, these outlier dimensions disrupt symmetric absmax quantization and favor asymmetric zeropoint quantization. This explains the results in our validation perplexity analysis. These observations appear to be universal as they occur for models trained in different software frameworks (fairseq, OpenAI, Tensorflow-mesh), and they occur in different inference frameworks (fairseq, Hugging Face Transformers). These outliers also appear robust to slight variations of the transformer architecture (rotary embeddings, embedding norm, residual scaling, different initializations). }
\label{tbl:outliers}
\resizebox{\textwidth}{!}{%
\begin{tabular}{lrlccccccc}\toprule
 &
  \multicolumn{1}{l}{} &
   
   &
  \multicolumn{2}{c}{Outliers} &
  \multicolumn{2}{c}{Frequency} &
   &
  \multicolumn{2}{c}{Top-1 softmax p} \\ \cmidrule{4-7} \cmidrule{9-10}  
Model &
  \multicolumn{1}{l}{PPL\bf{$\downarrow$}} &
  Params &
  Count &
  1-sided &
  Layers &
  SDims &
  Quartiles &
  w/ Outlier &
  No Outlier \\\toprule
GPT2  & 33.5 & 117M   & 1 & 1 & 25\%  & 6\%  & (-8, -7, -6)            & 45\% & 19\% \\
GPT2  & 26.0   & 345M   & 2 & 1 & 29\%  & 18\% & (6, 7, 8)                & 45\% & 19\% \\
FSEQ  & 25.7 & 125M  & 2 & 2 & 25\%  & 22\% & (-40, -23, -11)            & 32\%  & 24\% \\
GPT2  & 22.6 & 762M   & 2 & 0 & 31\%  & 16\% & (-9, -6, 9)               & 41\% & 18\% \\
GPT2  & 21.0   & 1.5B   & 2 & 1 & 41\%  & 35\% & (-11, -9, -7)    & 41\% & 25\% \\
FSEQ  & 15.9 & 1.3B  & 4 & 3 & 64\%  & 47\% & (-33, -21, -11) & 39\% & 15\% \\
FSEQ  & 14.4 & 2.7B  & 5 & 5 & 52\%  & 18\% & (-25, -16, -9)            & 45\% & 13\% \\
GPT-J & 13.8 & 6.0B   & 6 & 6 & 62\%  & 28\% & (-21, -17, -14)           & 55\% & 10\% \\\midrule
FSEQ  & 13.3 & 6.7B  & 6 & 6 & 100\% & 75\% & (-44, -40, -35)           & 35\% & 13\% \\
FSEQ  & 12.5 & 13B   & 7 & 6 & 100\% & 73\% & (-63, -58, -45)   & 37\% & 16\%\\\bottomrule
\end{tabular}%
}
\end{table}

\section{Inference Speedups and Slowdowns}
\label{app:inference}

\subsection{Matrix Multiplication benchmarks}

While our work focuses on memory efficiency to make models accessible, Int8 methods are also often used to accelerate inference. We find that the quantization and decomposition overhead is significant, and Int8 matrix multiplication itself only yields an advantage if the entire GPU is well saturated, which is only true for large matrix multiplication. This occurs only in LLMs with a model dimension of 4096 or larger.

Detailed benchmarks of raw matrix multiplication and quantization overheads are seen in Table~\ref{tbl:layer_speed}. We see that raw Int8 matrix multiplication in cuBLASLt begins to be two times faster than cuBLAS at a model size of 5140 (hidden size 20560). If inputs need to be quantized and outputs dequantized -- a strict requirement if not the entire transformer is done in Int8 -- then the speedups compared to 16-bit is reduced to 1.6x at a model size of 5140. Models with model size 2560 or smaller are slowed down. Adding mixed precision decomposition slows inference further so that only the 13B and 175B models have speedups.

These numbers could be improved significantly with optimized CUDA kernels for the mixed precision decomposition. However, we also see that existing custom CUDA kernels are much faster than when we use default PyTorch and NVIDIA-provided kernels for quantization which slow down all matrix multiplications except for a 175B model.

\begin{table}[h]
\centering
\caption{Inference speedups compared to 16-bit matrix multiplication for the first hidden layer in the feed-forward of differently sized GPT-3 transformers. The hidden dimension is 4x the model dimension. The 8-bit without overhead speedups assumes that no quantization or dequantization is performed. Numbers small than 1.0x represent slowdowns. Int8 matrix multiplication speeds up inference only for models with large model and hidden dimensions. }
\label{tbl:layer_speed}
\resizebox{\textwidth}{!}{%
\begin{tabular}{lllllllll}\toprule
GPT-3 Size                & Small & Medium & Large & XL    & 2.7B  & 6.7B  & 13B   & 175B  \\
Model dimension                 & 768   & 1024   & 1536  & 2048  & 2560  & 4096  & 5140  & 12288 \\\toprule
FP16-bit baseline           & 1.00x & 1.00x  & 1.00x & 1.00x & 1.00x & 1.00x & 1.00x & 1.00x \\
Int8 without overhead    & 0.99x & 1.08x  & 1.43x & 1.61x & 1.63x & 1.67x & 2.13x & 2.29x \\\midrule
Absmax PyTorch+NVIDIA     & 0.25x & 0.24x  & 0.36x & 0.45x & 0.53x & 0.70x & 0.96x & 1.50x \\
Vector-wise PyTorch+NVIDIA        & 0.21x & 0.22x  & 0.33x & 0.41x & 0.50x & 0.65x & 0.91x & 1.50x \\
Vector-wise & \textbf{0.43x} & \textbf{0.49x} & \textbf{0.74x} & \textbf{0.91x} & \textbf{0.94x} & \textbf{1.18x} & \textbf{1.59x} & \textbf{2.00x} \\
LLM.int8() (vector-wise+decomp) & 0.14x & 0.20x  & 0.36x & 0.51x & 0.64x & 0.86x & 1.22x & 1.81x\\\bottomrule
\end{tabular}%
}
\end{table}

\subsection{End-to-end benchmarks}

Besides matrix multiplication benchmarks, we also test the end-to-end inference speed of BLOOM-176B in Hugging Face. Hugging Face uses an optimized implementation with cached attention values. Since this type of inference is distributed and, as such, communication dependent, we expect the overall speedup and slowdown due to Int8 inference to be smaller since a large part of the overall inference runtime is the fixed communication overhead. 

We benchmark vs. 16-bit and try settings that use a larger batch size or fewer GPUs in the case of Int8 inference, since we can fit the larger model on fewer devices.
We can see results for our benchmark in Table~\ref{tbl:memory_footprint}. Overall Int8 inference is slightly slower but close to the millisecond latency per token compared to 16-bit inference.

\begin{table}[h]
\centering
\caption{Ablation study on the number of GPUs used to run several types of inferences of BLOOM-176B model. We compare the number of GPUs used by our quantized BLOOM-176B model together with the native BLOOM-176B model. We also report the \textit{per-token} generation speed in milliseconds for different batch sizes. We use our method integrated into transformers\citep{wolf2019huggingface} powered by accelerate library from HuggingFace to deal with multi-GPU inference. Our method reaches a similar performance to the native model by fitting into fewer GPUs than the native model.}
\label{tbl:memory_footprint}
\begin{tabular}{llccc}\toprule
Batch Size              & Hardware  & 1 & 8 & 32  \\\toprule
bfloat16 baseline           & 8xA100 80GB & \textbf{239} & \textbf{32}  & 9.94  \\\midrule
LLM.int8()     & 8xA100 80GB    & 253 & 34  & 10.44  \\
LLM.int8()       & 4xA100 80GB    & 246 & 33  & 9.40  \\
LLM.int8()      & 3xA100 80GB    & 247 & 33  & \textbf{9.11} \\\bottomrule
\end{tabular}
\end{table}

\section{Training Results}
\label{appendix:training}

We test Int8 training on a variety of training settings and compare to 32-bit baselines. We test separate settings for running the transformer with 8-bit feed-forward networks with and without 8-bit linear projections in the attention layer, as well at the attention iteself in 8-bit and compare against 32-bit performance. 
We test two tasks (1) language modeling on part of the RoBERTa corpus including Books~\citep{zhu2015aligning}, CC-News \citep{nagel2016cc}, OpenWebText \citep{gokaslan2019openwebtext}, and CC-Stories \citep{trinh2018simple}; and (2) neural machine translation (NMT) \citep{ott2018scaling} on WMT14+WMT16 \citep{machavcek2014results, sennrich2016edinburgh}. 

The results are shown in Table~\ref{tbl:traininglm} and Table~\ref{tbl:nmt}. We can see that for training, using the attention linear projections with Int8 data types and vector-wise quantization leads to degradation for NMT and for 1.1B language model but not for 209M language modeling. The results improve slightly if mixed-precision decomposition is used but is not sufficient to recover full performance in most cases. 
These suggests that training with 8-bit FFN layers is straightforward while other layers require additional techniques or different data types than Int8 to do 8-bit training at scale without performance degradation.

\begin{table}[h]
\centering
\caption{Initial results on small and large-scale language modeling. Doing attention in 8-bit severely degrades performance and performance cannot fully recovered with mixed-precision decomposition. While small-scale language models is close to baseline performance for both 8-bit FFN and 8-bit linear projects in the attention layers performance degrades at the large scale.}
\label{tbl:traininglm}
\begin{tabular}{cccccc}

& \multicolumn{3}{c}{Is 8-bit} & & \\\cmidrule{2-4}
Params & FFN & Linear & Attention & Decomp & PPL   \\\midrule
209M   &   &       &           & 0\%             & 16.74 \\
209M   & \checkmark   &      &           & 0\%             & 16.77 \\
209M   & \checkmark    & \checkmark .      &           & 0\%             & 16.83 \\\midrule
209M   & \checkmark    & \checkmark       &           & 2\%             & 16.78 \\
209M   & \checkmark    & \checkmark       &           & 5\%             & 16.77 \\
209M   & \checkmark    & \checkmark       &           & 10\%            & 16.80 \\\midrule
209M   & \checkmark    & \checkmark       & \checkmark          & 2\%             & 24.33 \\
209M   & \checkmark    & \checkmark       & \checkmark          & 5\%             & 20.00 \\
209M   & \checkmark    & \checkmark       & \checkmark          & 10\%            & 19.00 \\\midrule
1.1B   &    &      &           & 0\%             & 9.99  \\
1.1B   & \checkmark    &      &           & 0\%             & 9.93  \\
1.1B   & \checkmark    & \checkmark       &           & 0\%             & 10.52 \\
1.1B   & \checkmark    & \checkmark       &           & 1\%             & 10.41\\\bottomrule
\end{tabular}%
\end{table}

\begin{table}[]
\centering
\caption{Neural machine translation results for 8-bit FFN and linear attention layers for WMT14+16. Decomp indicates the percentage that is computed in 16-bit instead of 8-bit. The BLEU score is the median of three random seeds.}
\label{tbl:nmt}
\begin{tabular}{cccc}\toprule
\multicolumn{2}{c}{Is 8-bit} & & \\\cmidrule{1-2}
\multicolumn{1}{l}{FFN} & \multicolumn{1}{l}{Linear} & \multicolumn{1}{l}{Decomp} & \multicolumn{1}{l}{BLEU} \\\midrule
 &  &  0\% & 28.9                         \\
\checkmark &  & 0\% & 28.8                         \\
\checkmark & \checkmark & 0\%  & \multicolumn{1}{l}{unstable} \\
\checkmark & \checkmark & 2\%  & 28.0                           \\
\checkmark & \checkmark & 5\%  & 27.6                         \\
\checkmark & \checkmark & 10\% & 27.5   \\\bottomrule                     
\end{tabular}%
\end{table}

\section{Fine-tuning Results}

We also test 8-bit finetuning on RoBERTa-large finetuned on GLUE. We run two different setups: (1) we compare with other Int8 methods, and (2) we compare degradation of finetuning with 8-bit FFN layers as well as 8-bit attention projection layers comparel to 32-bit. We finetune with 5 random seeds and report median performance. 

Table~\ref{tbl:finetune} compares with different previous 8-bit methods for finetuning and shows that vector-wise quantization improves on other methods. 
Table~\ref{tbl:breakdownglue} shows the performance of FFN and/or linear attention projections in 8-bit as well as improvements if mixed-precision decomposition is used. We find that 8-bit FFN layers lead to no degradation while 8-bit attention linear projections lead to degradation if not combined with mixed-precision decomposition where at least the top 2\% magnitude dimensions are computed in 16-bit instead of 8-bit.
These results highlight the critical role of mixed-precision decomposition for finetuning if one wants to not degrade performance.

\begin{table}[]
\centering
\caption{GLUE finetuning results for quantization methods for the feedforward layer in 8-bit while the rest is in 16-bit. No mixed-precision decomposition is used. We can see that vector-wise quantization improve upon the baselines.}
\label{tbl:finetune}
\resizebox{\textwidth}{!}{%
\begin{tabular}{lccccccccc}\toprule
Method             & MNLI & QNLI & QQP  & RTE           & SST-2 & MRPC & CoLA & STS-B & Mean  \\\midrule
32-bit Baseline    & 90.4 & 94.9 & 92.2 & 84.5          & 96.4  & 90.1 & 67.4 & 93.0    & 88.61 \\
32-bit Replication & 90.3 & 94.8 & 92.3 & 85.4          & 96.6  & 90.4 & 68.8 & 92.0    & 88.83 \\\midrule
Q-BERT \citep{shen2020q}    & 87.8 & 93.0   & 90.6 & 84.7          & 94.8  & 88.2 & 65.1 & 91.1  & 86.91 \\
Q8BERT  \citep{zafrir2019q8bert}           & 85.6 & 93.0   & 90.1 & 84.8          & 94.7  & 89.7 & 65.0   & 91.1  & 86.75 \\
PSQ   \citep{chen2020statistical}             & 89.9 & 94.5 & 92.0 & \textbf{86.8} & 96.2  & 90.4 & 67.5 & 91.9  & 88.65 \\\midrule
Vector-wise & \textbf{90.2} & \textbf{94.7} & \textbf{92.3} & 85.4 & \textbf{96.4} & \textbf{91.0} & \textbf{68.6} & \textbf{91.9} & \textbf{88.81}\\\bottomrule
\end{tabular}%
}
\end{table}

\begin{table}[h]
\centering
\caption{Breakdown for 8-bit feedforward network (FFN) and linear attention layers for GLUE. Scores are median of 5 random seeds. Decomp indicates the percentage that is decomposed into 16-bit matrix multplication. Compared to inference, fine-tuning appears to need a higher decomp percentage if the linear attention layers are also converted to 8-bit.}
\label{tbl:breakdownglue}
\resizebox{\textwidth}{!}{%
\begin{tabular}{ccccccccccccc}\toprule
\multicolumn{2}{c}{Is 8-bit} & & & & & & & & & &\\\cmidrule{1-2}
FFN & Linear & Decomp & MNLI  & QNLI  & QQP  & RTE  & SST-2 & MRPC  & CoLA  & STS-B & MEAN     \\\midrule
   &       &       0\%      & 90.4  & 94.9 & 92.2 & 84.5 & 96.4  & 90.1  & 67.4 & 93.0 & 88.6 \\
\checkmark   &       &     0\%        & 90.2  & 94.7  & 92.3 & 85.4 & 96.4  & 91.0 & 68.6  & 91.9 & 88.8 \\
\checkmark   & \checkmark      & 0\%             & 90.2 & 94.4  & 92.2 & 84.1 & 96.2 & 89.7  & 63.6 & 91.6 & 87.7 \\
\checkmark   & \checkmark      & 1\%             & 90.0    & 94.6  & 92.2 & 83.0   & 96.2  & 89.7  & 65.8  & 91.8  & 87.9  \\
\checkmark   & \checkmark      & 2\%             & 90.0    & 94.5  & 92.2 & 85.9 & 96.7  & 90.4  & 68.0    & 91.9  & 88.7     \\
\checkmark   & \checkmark     & 3\%            & 90.0    & 94.6  & 92.2 & 86.3 & 96.4  & 90.2  & 68.3  & 91.8  & 88.7  \\\bottomrule
\end{tabular}%
}
\end{table}

\end{document}